\RequirePackage[T1]{fontenc}
\documentclass[conference]{IEEEtran}
\usepackage{amsmath,amssymb,bm}
\usepackage{graphicx,booktabs,tabularx,array,cite}
\usepackage{algorithm,algpseudocode}
\usepackage[hidelinks]{hyperref}

\hypersetup{pdftitle={TOFS: Feature Selection with Swap Supervision for Semantic Communication over Fading Channels},pdfauthor={Zhiyi Zhu, Eiji Takimoto, Patrick Finnerty, Chikara Ohta},pdfsubject={VTC2027-Spring, Signal Processing for Wireless Communications}}

\newcommand{\CE}{\ell_{\mathrm{CE}}}
\newcommand{\sg}{\operatorname{sg}}
\title{\textit{TOFS}: Feature Selection with Swap Supervision for Semantic Communication over Fading Channels}
\author{
\IEEEauthorblockN{Zhiyi Zhu\textsuperscript{1,*}, Eiji Takimoto\textsuperscript{2},
Patrick Finnerty\textsuperscript{1}, Chikara Ohta\textsuperscript{1}}
\IEEEauthorblockA{\textsuperscript{1}Graduate School of System Informatics, Kobe University, Kobe, Japan}
\IEEEauthorblockA{\textsuperscript{2}Information Technology Center, Nara Women's University, Nara, Japan}
\IEEEauthorblockA{* Corresponding author email: syu@port.kobe-u.ac.jp}}

\begin{document}
\maketitle
\begin{abstract}
Task-oriented semantic communication transmits task-relevant features to support receiver inference under limited communication resources and adverse channel conditions. For image classification, the usefulness of a feature depends on the other transmitted features and the distortion introduced by the wireless channel. Effective feature prioritization should therefore account for classification performance after transmission of the selected features. This paper proposes Task-Oriented Feature Selection (\textit{TOFS}) framework and method, which evaluates feature replacements using receiver classification loss. Loss differences confirmed under independent transmission samples provide pairwise ranking supervision for a selector network. The trained selector predicts priorities in a single forward pass. CIFAR-10 simulations demonstrate improved classification accuracy over an importance-based feature selection baseline at the same transmitted feature count and average transmit power. The proposed \textit{TOFS} framework shows 84\% accuracy performance under Rayleigh fading and Gaussian noise, compared with the 68\% performance of a SOTA method.
\end{abstract}

\section{Introduction}
Task-oriented semantic communication aims to support receiver tasks by transmitting task-relevant information \cite{shao2022}. In remote image classification, this information is represented by features extracted at the transmitter. Limited communication resources motivate transmitting only selected features, while fading and noise can corrupt their values. Moreover, transmitting more features does not necessarily improve classification accuracy \cite{liu2026}. Reliable classification therefore requires identifying features that remain useful after selection and transmission.

Deep joint source and channel coding (JSCC) learns image representations for robust reconstruction over noisy and fading channels \cite{bourtsoulatze2019}. For receiver inference, Shao \emph{et al.} combine an information bottleneck with variable-length encoding to balance inference accuracy and communication cost \cite{shao2022}. Adaptable semantic compression further improves task performance through joint compression and resource allocation \cite{liu2024asc}.

Importance-based transmission further distinguishes semantic content according to its contribution to communication tasks. Cai \emph{et al.} select task-relevant scene graphs and allocate resources to reduce energy consumption under delay constraints \cite{cai2025}. Qiao \emph{et al.} assign important semantic entities to reliable polarized channels to improve semantic preservation \cite{qiao2025}. Liu \emph{et al.} combine message importance with feature priorities predicted from classification loss variation, improving classification under harsh channel conditions \cite{liu2026}.

These studies demonstrate the benefits of selecting or protecting important content. A remaining issue is how to evaluate each feature's contribution within a jointly transmitted feature set. 
Classification depends on the features retained together, while fading and noise distort their values. When power normalization is shared, changing the selected set can also change restoration errors. This motivates the question: \emph{how should features be prioritized when their contribution to classification depends on the jointly transmitted feature set and channel distortion?}

The main contributions of this paper are as follows.
\begin{itemize}
 \item We propose Task-Oriented Feature Selection (\textit{TOFS}), a framework that learns feature priorities using receiver classification losses of jointly transmitted feature sets. 
 Evaluation after transmission accounts for both the retained classification information and channel distortion.
 \item We develop a training procedure that converts classification loss differences after feature replacements, confirmed under independent transmission samples, into pairwise ranking supervision. The resulting selector predicts priorities in one forward pass.
 \item CIFAR-10 simulations at $-2$ dB with 100 transmitted features demonstrate improved classification over predicted loss variation and a matched straight-through training control at equal average transmit power. The analysis shows that a smaller normalization scale reduces restoration errors of transmitted features under the adopted receiver model.
\end{itemize}

\section{System Model and Problem Formulation}
\begin{figure*}[t]
 \centering\includegraphics[width=\textwidth]{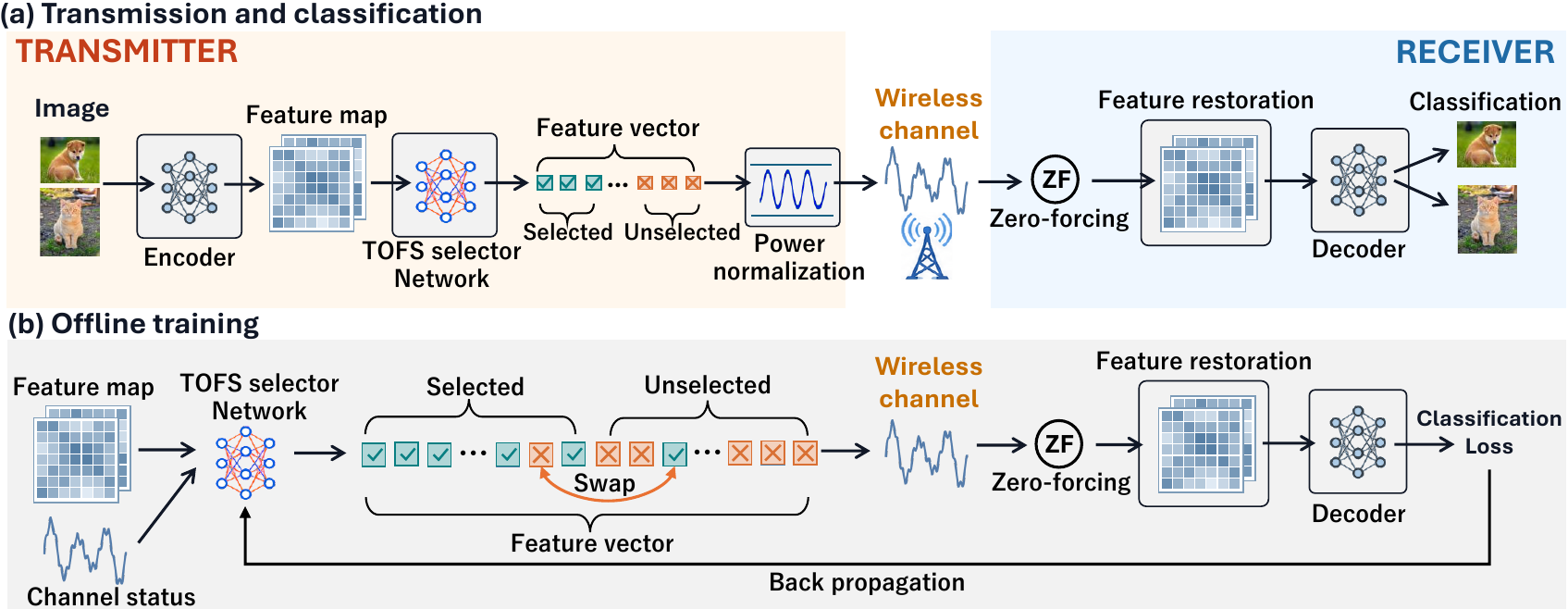}
 \caption{Overview of the proposed \textit{TOFS} framework and feature selection training.}
 \label{fig:workflow}
\end{figure*}
\subsection{System Architecture}
Fig.~\ref{fig:workflow}(a) shows a transmitter for feature extraction and selection and a receiver for image classification. The encoder produces a spatial feature map, which is pooled into a feature vector of scalar values. Given this map and a fixed link descriptor of the reference transmission configuration, the \textit{TOFS} selector network predicts one priority score per feature value. Feature selection gathers the highest-priority original values and omits unselected values. Power normalization scales these values to a common average transmit power for transmission over the wireless channel.

At the receiver, zero-forcing (ZF) equalization uses known channel gains to compensate fading. Feature restoration reverses transmit scaling, places recovered values at their original coordinates, and inserts zeros at omitted positions. The decoder outputs class scores from this restored vector; the largest score determines the predicted class.

Offline training in Fig.~\ref{fig:workflow}(b) uses independently confirmed classification loss differences between transmitted feature sets, with fixed encoder and decoder. Online inference requires one forward pass through the trained selector network.

For image $x$, let $\bm z$ be the feature map and $\bm f\in\mathbb R^D$ the vector of $D$ feature values. Coordinate $i$ indexes scalar feature value $f_i$. The selector outputs $\bm s_\theta=P_\theta(\bm z,\bm c)\in\mathbb R^D$, with parameters $\theta$ and fixed descriptor $\bm c$. Let $K$ denote the transmitted feature count, with $1\le K\le D$; the top $K$ scores define the selected feature coordinate set $S_\theta(x)$. 
At $K=D$, all feature values are transmitted. The restored vector $\widehat{\bm f}$ enters the linear decoder $G(\widehat{\bm f})=W\widehat{\bm f}+\bm b$, with weights $W$ and bias $\bm b$.

\subsection{Wireless Channel Model}
In the baseband model, each normalized feature value is a continuous real-valued channel input, termed a symbol \cite{bourtsoulatze2019}. Sending one amplitude constitutes one channel use. A random permutation $\pi$ maps sorted selected indices to channel uses independently of scores and gains. 
For feature set $S$ of size $K$, the root mean square (RMS) scale $a_S$ normalizes selected value $f_{i_j}$ assigned to channel use $j$:
\begin{equation}
 a_S=\sqrt{\frac{1}{K}\sum_{i\in S}f_i^2},\qquad u_j=\frac{f_{i_j}}{a_S},\qquad
 \frac{1}{K}\sum_{j=1}^{K}u_j^2=1.
 \label{eq:normalize}
\end{equation}
For nonzero feature sets, normalization gives unit average transmit power per channel use and normalized energy $K$. Equal feature counts therefore imply equal feature transmission energy. A numerical floor handles all-zero sets, and receiver multiplication by $a_S$ restores feature scale.

After ideal phase compensation, Rayleigh fading and additive Gaussian noise yield the channel and restored feature values
\begin{equation}
 r_j=h_ju_j+\frac{z_j}{\sqrt{\rho}},\qquad
 \widehat f_{i_j}=f_{i_j}+a_S\frac{z_j}{\sqrt{\rho}\,h_j},
 \label{eq:channel}
\end{equation}
where $r_j$ is the received sample, $\gamma_{\mathrm{dB}}$ is nominal average signal-to-noise ratio (SNR), and $\rho=10^{\gamma_{\mathrm{dB}}/10}$. Noise samples satisfy $z_j\sim\mathcal N(0,1)$, and independent Rayleigh gains satisfy $\mathbb E[h_j^2]=1$. Noise variance is $1/\rho$ at unit average transmit power, with SNR defined per real dimension. Perfect receiver channel state information (CSI) supplies $h_j$; the selector receives no instantaneous gains. Dividing by $h_j$ compensates attenuation and amplifies noise; the restoration denominator uses $\max(h_j,10^{-4})$ for numerical stability.

The receiver knows indices, permutation, and scale exactly. Unselected features require no channel uses, and their decoder coordinates remain zero; selected zero values still use the channel and acquire noise. In clean transmission, selected values are recovered exactly without fading or noise, while untransmitted coordinates remain zero. A transmission instance comprises gains, noise, and a permutation.

\subsection{Feature Selection Objective}
For selected set $S$ and transmission instance $\xi$, receiver cross-entropy is $\CE(G(\widehat{\bm f}(S;\xi)),y)$, with true class $y$. The loss evaluates classification of the entire restored vector, including recovered selected values and zeros elsewhere. Loss differences supervise \textit{TOFS}; accuracy measures task performance. Training comparisons preserve $K$ to isolate feature choice; Section~\ref{sec:design} examines different counts.

\section{Proposed \textit{TOFS} Method}
\subsection{Feature Scoring and Training Overview}
Fig.~\ref{fig:workflow}(b) and Algorithm~\ref{alg:tofs} describe training. The current set and swaps block generates replacement candidates. Transmission evaluation compares current and candidate sets using receiver classification loss and training labels. Independent confirmation checks preferences under new transmission samples; the network update trains only the selector network using ranking loss.

\begin{algorithm}[H]
\caption{\textit{TOFS} training and inference}
\label{alg:tofs}
\begin{algorithmic}[1]
\Require Labeled training images, fixed encoder and decoder, transmitted feature count $K$, initialized selector network
\For{each minibatch over the training epochs}
 \State Encode images; score features and select top $K$ coordinates
 \State Form candidates by swapping a selected and an unselected coordinate
 \State Sample independent search and confirmation banks
 \State Evaluate current and candidate sets in both banks
 \State Accept matching signs and $|\Delta_{pq}^{(b)}|>\eta$ in both banks
 \State Update selector network with ranking loss if targets exist
\EndFor
\Statex \textbf{Inference:} encode image, score and select features, normalize and transmit; restore features and classify
\end{algorithmic}
\end{algorithm}

\subsection{Candidate Replacements and Confirmation}
Replacement training uses $K<D$ so that unselected features are available. For selected coordinate $p\in S$ and unselected coordinate $q\notin S$, a candidate is
\begin{equation}
 S_{p\rightarrow q}=(S\setminus\{p\})\cup\{q\}.
 \label{eq:swap}
\end{equation}
Each candidate preserves the number of transmitted feature values. The removed coordinate becomes zero at the decoder, and the added coordinate receives a recovered value. Candidates remove coordinates with low scores and add unselected coordinates with high scores or random samples. The incoming feature uses the outgoing feature's channel use; retained assignments stay fixed and normalization is recomputed. The current set and candidate share gains and noise, making their loss difference a paired comparison.

Two independent banks of 32 transmission instances measure and confirm preferences. Let $\ell_t^{(b)}(A)$ denote receiver cross-entropy for set $A$ under instance $t$ in bank $b$:
\begin{equation}
 \Delta_{pq}^{(b)}=\frac{1}{32}\sum_{t=1}^{32}
 \left[\ell_t^{(b)}(S_{p\rightarrow q})-\ell_t^{(b)}(S)\right].
 \label{eq:delta}
\end{equation}
Negative changes favor the incoming feature; positive changes favor the outgoing feature. The confirmation bank repeats the comparison under new transmission samples. Matching signs and magnitudes above $\eta$ in both banks establish a ranking target; other pairs receive zero weight.

\subsection{Ranking Optimization and Inference}
A pairwise ranking loss learns the confirmed preferences \cite{burges2005}. Define $d_{pq}=-\operatorname{sign}(\Delta_{pq}^{(1)}+\Delta_{pq}^{(2)})$ and $w_{pq}=\min\{|\Delta_{pq}^{(1)}|,|\Delta_{pq}^{(2)}|,1\}$, with $w_{pq}=0$ for rejected pairs. Let $\widetilde s_i=s_i/\sg(\max\{\operatorname{std}(\bm s),0.05\})$, where $\sg$ blocks gradients. The update minimizes
\begin{equation}
 L_{\rm rank}=\frac{\sum_{p,q}w_{pq}\log\!\left(1+
 \exp[-d_{pq}(\widetilde s_q-\widetilde s_p)/\tau]\right)}
 {\sum_{p,q}w_{pq}},
 \label{eq:rank}
\end{equation}
with temperature $\tau$ and minibatch normalization over accepted pairs. Loss differences supply detached preference labels and weights; backpropagation through scores updates only the selector network using~\eqref{eq:rank}.

Inference uses one selector pass to gather original values at the top $K$ scored coordinates.

\section{Simulation Results and Discussion}
\label{sec:results}
\subsection{Simulation Setup}
Experiments use CIFAR-10 \cite{krizhevsky2009} with 45,000 training, 5,000 validation, and all 10,000 test images, resized to $224\times224$. A ResNet-18 encoder \cite{he2016} produces a $512\times7\times7$ intermediate map and $D=512$ pooled features. The linear decoder has ten outputs. The final rectified linear unit (ReLU) gives nonnegative pooled features and transmit amplitudes; noisy restored features can take either sign. ImageNet initialization precedes clean task training; the encoder and decoder are then frozen.

The selector network uses a residual block, two $1\times1$ convolutions, pooling, and an image context with 128 dimensions. Fused contexts and internally predicted class logits feed the scoring head. Ties use index order. The descriptor contains 104 values describing nominal SNR, selected feature fraction $K/D$, channel type, and effective SNRs from Rayleigh quantiles. Its reference ($K=100$, $-2$ dB, Rayleigh) stays fixed; the SNR sweep changes the channel with fixed scores and selections.

\begin{table}[t]
\caption{Simulation parameters.}
\label{tab:setup}
\centering\small
\begin{tabularx}{\columnwidth}{@{}lX@{}}
\toprule
Item & Setting \\
\midrule
Link and receiver & $K=100$, Rayleigh + noise, $-2$ dB; ZF \\
Continuation & 30 epochs, Adam, batch 256; learning rate $10^{-3}$ with cosine decay; weight decay $10^{-5}$ \\
Swap feedback & $8\times(8+8)=128$ pairs; $32+32$ independent transmission instances; $\eta=10^{-3}$, $\tau=0.35$ \\
Test randomness & Channel seeds 2026--2028; permutation seed 7202026 \\
SNR sweep & $\{-10,-6,-2,0,2,6,10\}$ dB \\
\bottomrule
\end{tabularx}
\end{table}

\begin{figure}[t]
 \centering\includegraphics[width=\columnwidth]{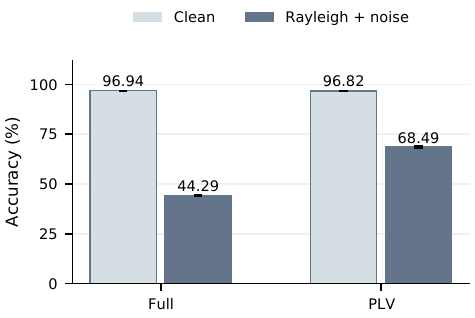}
 \caption{Classification accuracy under clean and impaired transmission.}
 \label{fig:motivation}
\end{figure}

We compare \textit{Full}, \textit{Random}, \textit{PLV} (predicted loss variation), and \textit{TOFS}. \textit{Full} sends all features; \textit{Random} selects 100 coordinates uniformly; \textit{PLV} and \textit{TOFS} send values at their 100 highest-scoring coordinates. The reference count follows \cite{liu2026}; equal counts isolate selection effects at equal feature transmission energy. Methods sending 100 features share gains, noise, and permutations. \textit{Full} uses more channel uses and energy with the same channel distributions. Clean and impaired tests keep identical selected coordinates. Accuracy averages three channel seeds; $\pm$ and error bars denote their sample standard deviation.

The reference \textit{PLV} adapts the feature ranking component of \cite{liu2026} to the common system. Its network predicts the loss increase from removing one coordinate of the complete clean representation:
\begin{equation}
 v_i=\CE\bigl(G(\bm f-f_i\bm e_i),y\bigr)-\CE\bigl(G(\bm f),y\bigr),
 \label{eq:liu}
\end{equation}
where $\bm e_i$ is a coordinate basis vector. Mean squared error (MSE) fits scores to these targets. \textit{PLV} removes one feature with all others present and undistorted; \textit{TOFS} evaluates the entire restored vector with untransmitted coordinates set to zero.

A straight-through estimator (STE) approximates gradients through discrete top $K$ selection \cite{bengio2013}. \textit{TOFS} and its STE control start from one network trained for 30 epochs using impaired cross-entropy and three draws per image. They continue for 30 epochs with 64 draws per image. STE retains its sigmoid derivative and gradient path through normalization and channel distortion; \textit{TOFS} uses~\eqref{eq:rank}.

Training pairs eight lowest-scoring selected coordinates with eight highest-scoring and eight random unselected coordinates (128 candidates). Each image requires 8192 candidate evaluations plus baselines, adding computation beyond the matched STE setting.

Checkpoint selection uses validation accuracy for the task model, validation MSE for \textit{PLV}, and the predeclared final epoch for \textit{TOFS} and its STE control. Repeated use of the test set makes this evaluation retrospective. Results from confirmed replacements use CIFAR-10; fresh images and further datasets would assess independent performance and transfer.

\subsection{Feature Selection and Channel Distortion}
Clean tests retain selected values exactly and set untransmitted coordinates to zero. In Fig.~\ref{fig:motivation}, \textit{PLV} retains accuracy close to \textit{Full}, yet outperforms it under Rayleigh fading and noise at $-2$ dB. \textit{Random} has lower clean accuracy than \textit{PLV} at the same feature count (Table~\ref{tab:accuracy}). These comparisons motivate evaluating both the selected representation and the additional channel distortion.

\subsection{Classification Performance}
\begin{table}[t]
\caption{Classification accuracy.}
\label{tab:accuracy}
\centering\small
\begin{tabular}{@{}lrr@{}}
\toprule
Method & Clean (\%) & Rayleigh (\%) \\
\midrule
\textit{Full} & 96.94 & $44.29 \pm 0.29$ \\
\textit{Random} & 73.09 & $24.16 \pm 0.16$ \\
\textit{PLV} & 96.82 & $68.49 \pm 0.35$ \\
\textit{TOFS} & 96.81 & $84.17 \pm 0.37$ \\
\bottomrule
\end{tabular}
\end{table}
Table~\ref{tab:accuracy} shows that \textit{TOFS} improves accuracy over \textit{PLV} by 15.68 percentage points at $-2$ dB. Their clean accuracies differ by only 0.01 percentage points, indicating similarly high classification performance with untransmitted coordinates set to zero.

\begin{figure}[t]
 \centering\includegraphics[width=\columnwidth]{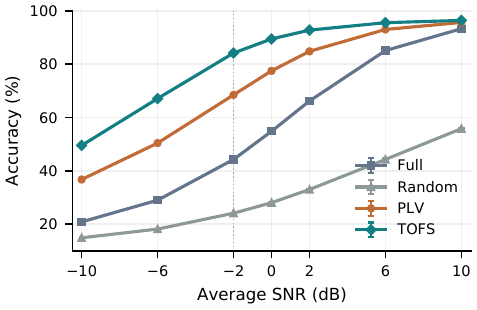}
 \caption{Classification accuracy versus average SNR.}
 \label{fig:snr}
\end{figure}
As shown in Fig.~\ref{fig:snr}, \textit{TOFS} achieves higher accuracy than \textit{PLV} across the seven SNRs with a fixed selection rule. At $-2$ dB, matched STE reaches 73.34\% impaired accuracy and 96.71\% clean accuracy, versus 84.17\% and 96.81\% for \textit{TOFS}. The gain of 10.82 percentage points demonstrates the benefit of the proposed training procedure under matched initialization, training duration, and channel draws.

\subsection{Selected Features and Restoration Error}
\begin{figure}[t]
 \centering\includegraphics[width=\columnwidth]{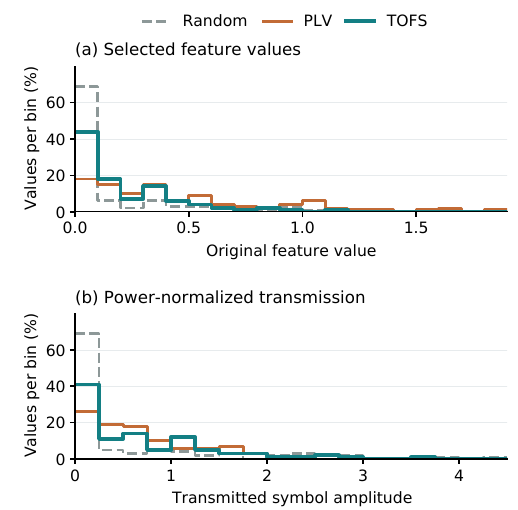}
 \caption{Distributions of selected values and transmit amplitudes for one image.}
 \label{fig:selection-example}
\end{figure}
Fig.~\ref{fig:selection-example} compares 100 selected values and normalized transmit amplitudes for test image 0, preselected by index. Histograms use common bin widths of 0.1 and 0.25 and report the percentage of selected values in each bin.

\textit{TOFS} selects 22 values below 0.01, compared with one for \textit{PLV}. Their RMS scales are 0.3167 and 0.6264, and their sets share 57 coordinates. Both sets of normalized transmit amplitudes have mean square one. Selecting more small values therefore changes how transmit power is distributed among retained coordinates.

\begin{figure*}[t]
 \centering\includegraphics[width=\textwidth]{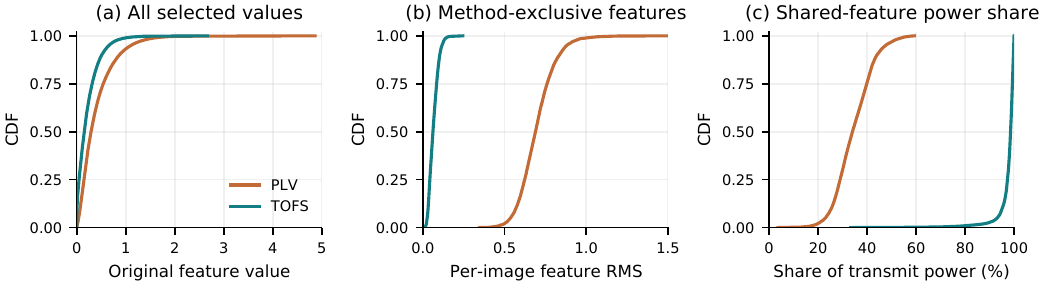}
 \caption{Selected feature amplitudes and distribution of transmit power.}
 \label{fig:distributions}
\end{figure*}
The cumulative distribution functions (CDFs) in Fig.~\ref{fig:distributions} show that \textit{TOFS} selects smaller feature values than \textit{PLV}. Panel (a) covers one million original selected values per method ($10{,}000\times100$). \textit{TOFS}'s median is 0.1510 versus 0.2912 for \textit{PLV}; 13.58\% versus 1.37\% are below 0.01.

Let $S_P$ and $S_T$ denote the sets selected by \textit{PLV} and \textit{TOFS}, respectively. Their shared coordinates form $C=S_P\cap S_T$, and $U_m=S_m\setminus C$ contains coordinates exclusive to method $m$. Mean overlap is 62.77, leaving 37.23 exclusive coordinates per method. Panel (b) isolates these differences using the exclusive RMS $r_{U_m}=\sqrt{|U_m|^{-1}\sum_{i\in U_m}f_i^2}$, which measures their overall amplitude. Means across images are 0.6965 for \textit{PLV} and 0.0651 for \textit{TOFS}. Shared coordinates retain identical original values, while their share of transmit power depends on the entire selected set:
\begin{equation}
 p_{C,m}=\frac{\sum_{i\in C}(f_i/a_{S_m})^2}
 {\sum_{i\in S_m}(f_i/a_{S_m})^2}
 =\frac{\sum_{i\in C}f_i^2}{\sum_{i\in S_m}f_i^2}.
 \label{eq:shared_power}
\end{equation}
Smaller exclusive values increase the shared features' power share under common normalization. Panel (c) gives mean shares of 97.61\% for \textit{TOFS} and 34.52\% for \textit{PLV} at equal energy for the transmitted features. Identical features thus receive different power shares. Panels (b,c) show one statistic per test image for the final sets; \textit{TOFS} training starts from STE.

For an individual replacement, the normalization scale obeys
\begin{equation}
 a_{S_{p\rightarrow q}}^2=a_S^2+\frac{f_q^2-f_p^2}{K}.
 \label{eq:scale_swap}
\end{equation}
If $|f_q|<|f_p|$, the scale decreases. A retained feature assigned to the same channel use then satisfies
\begin{equation}
 e_j=\widehat f_{i_j}-f_{i_j}=a_S\epsilon_j,\quad
 \epsilon_j=\frac{z_j}{\sqrt{\rho}h_j},\quad
 e'_j=\frac{a_{S_{p\rightarrow q}}}{a_S}e_j.
 \label{eq:error}
\end{equation}
Equation~\eqref{eq:error} shows how a smaller scale reduces retained errors under identical gains and noise. Mean $a_S$ is 0.5225 for \textit{PLV} and 0.3072 for \textit{TOFS}, which has a smaller scale for every image. At $-2$ dB, root mean squared error (RMSE) is computed over 100 transmitted coordinates per image and channel seed. Averaging the 30,000 RMSEs gives 1.7976 and 1.0569, respectively. Errors indexed by channel use and divided by $a_S$ coincide under shared draws; shared features can occupy different channel uses.

This RMSE covers transmitted coordinates; the classification effect of leaving unselected coordinates at zero is evaluated through receiver loss and accuracy. Classification feedback thus assesses the selected feature contributions jointly with transmission errors scaled by $a_S$.

\begin{figure}[t]
 \centering\includegraphics[width=\columnwidth]{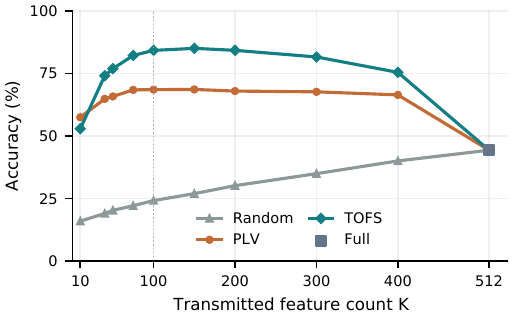}
 \caption{Classification accuracy versus feature count.}
 \label{fig:budget}
\end{figure}
\subsection{Analysis of Feature Count and Candidate Policies}
\label{sec:design}
\emph{Feature count.} For the trained \textit{TOFS} selector, accuracy does not increase monotonically with the number of transmitted features. We evaluate $K=10,40,50,75,100,150,200,300,400,512$ over seven SNRs with frozen models, scores, and descriptor. Validation (5,000 images) and test results are separate. Fig.~\ref{fig:budget} averages 10,000 test images and three channel seeds at $-2$ dB. At 512, all methods send the complete vector; channel uses and transmit energy scale with $K$.

\textit{TOFS} peaks at 84.99\% at $K=150$, compared with 84.17\% at $K=100$, and falls to 44.29\% when all 512 features are transmitted. Validation accuracy is 84.62\% and 84.57\% at $K=100$ and 200, respectively. The reference count performs close to the best evaluated count at $-2$ dB. At $K=10$, \textit{PLV} outperforms \textit{TOFS}, achieving 57.44\% versus 52.84\%.

\begin{table}[t]
\caption{Candidate selection strategies.}
\label{tab:candidate-ablation}
\centering\small
\setlength{\tabcolsep}{4pt}
\begin{tabular}{@{}llrrr@{}}
\toprule
Out & In & $P$ & Acc. (\%) & $t/t_8$ \\
\midrule
B4 & T4+R4 & 32 & 84.00 & 0.90 \\
\textbf{B8} & \textbf{T8+R8} & 128 & 84.39 & 1.00 \\
B16 & T16+R16 & 512 & 84.20 & 1.43 \\
R8 & T8+R8 & 128 & 83.71 & 0.99 \\
T8 & T8+R8 & 128 & 65.38 & 0.99 \\
B8 & R16 & 128 & 84.25 & 0.99 \\
B8 & B8+R8 & 128 & 84.47 & 0.99 \\
\bottomrule
\end{tabular}
\end{table}
\emph{Candidate policy.} Seven training variants share initialization, 45,000 training images, 30 epochs, optimizer, and 64 draws at $K=100$ and $-2$ dB (Table~\ref{tab:candidate-ablation}). Out and In specify the removal and addition pools. B denotes lowest scores, T highest scores, and R uniform sampling; suffixes indicate pool sizes. $P$ counts proposals; $t/t_8$ gives relative training time. The default configuration achieves 84.39\% here; the main result of 84.17\% is from the original run. CUDA accumulation of ranking gradients can introduce variation between runs.

With 128 proposals, lowest-scoring, random, and highest-scoring removal pools yield 84.39\%, 83.71\%, and 65.38\% accuracy, favoring preservation of high-scoring features. Widths 4, 8, and 16 give similar accuracy; width 16 costs $1.43\times$ the training time. Paired bootstrap resampling of images gives 95\% intervals of $[-0.79,0.02]$ and $[-0.58,0.17]$ points for widths 4 minus 8 and 16 minus 8, respectively. These exploratory intervals quantify uncertainty from resampling images for these checkpoints.

\subsection{Discussion}
Clean accuracy assesses how well the selected representation, with zeros at untransmitted coordinates, supports the fixed classifier. The noisy results show that retaining clean accuracy alone is insufficient to predict performance after transmission. Replacement losses jointly evaluate the effects of zero-filled unselected coordinates, normalization, and channel distortion. The matched STE comparison establishes a gain for candidate search, confirmation, and ranking together; it does not isolate each component's contribution.

Practical transmission also requires indices, scale information, and pilots. Under a specified signaling scheme, $N_{\rm total}=K+N_{\rm side}+N_{\rm pilot}$ counts feature, side-information, and pilot channel uses. For $D=512$, $K=100$, indices require 900 raw bits as a list or 512 bits as a bitmap. \textit{PLV} and \textit{TOFS} share the nominal index format; \textit{Full} requires no index list. Quantization, header protection, coding, and pilots determine rate, latency, and energy.

Evaluation assumes ZF, exact side information and receiver CSI, independent Rayleigh gains, and one task model and initialization. Channel seeds quantify transmission variability. Further datasets, independent tests, receiver designs, correlated fading, gain estimation errors, index errors, and processing costs remain for future evaluation.

\section{Conclusion}
\textit{TOFS} learns feature priorities from independently confirmed replacement preferences for classification over fading channels. With 100 transmitted features at $-2$ dB, CIFAR-10 results demonstrate improved accuracy over \textit{PLV} and the matched STE control. The analysis shows how normalization scales restoration errors of transmitted values under the adopted ZF receiver.

\end{document}